%% file: main.tex
\documentclass{article} 
\usepackage{tccml_iclr2025_conference,times}

\input{math_commands.tex}

\usepackage{hyperref}
\usepackage{url}
\usepackage{mathtools}
\usepackage{amsmath, amssymb, amsfonts}
\usepackage{bbm}
\usepackage{booktabs}
\usepackage{xcolor}
\usepackage{cleveref}

\definecolor{darkerblue}{HTML}{07359A}
\definecolor{darkerred}{HTML}{801D02}

\usepackage{tikz}
\usepackage{pgfplots}
\usepgfplotslibrary{groupplots}
\pgfplotsset{compat=newest}

 \newcommand{\repo}[1]{\href{https://github.com/SohirMaskey/raincast-gnn}{#1}}
 
\title{Graph Neural Networks for Enhancing Ensemble Forecasts of Extreme Rainfall}

\author{{Christopher Bülte\thanks{ Corresponding author.} , Sohir Maskey, Philipp Scholl, Jonas von Berg} \\
  Ludwig-Maximilians-Universit\"at M\"unchen \\
  Munich Center for Machine Learning (MCML) \\
	Munich, Germany\\
	\texttt{\{buelte, maskey, scholl, berg\}@math.lmu.de} \\
	 \And  
    {Gitta Kutyniok}\\
  Ludwig-Maximilians-Universit\"at M\"unchen \\
      University of Troms\o{} \\
  DLR-German Aerospace Center \\
  Munich Center for Machine Learning (MCML) \\
	Munich, Germany\\
	\texttt{kutyniok@math.lmu.de} \\
}

\iclrfinalcopy 
\begin{document}

\maketitle

\begin{abstract}
Climate change is increasing the occurrence of extreme precipitation events, threatening infrastructure, agriculture, and public safety. Ensemble prediction systems provide probabilistic forecasts but exhibit biases and difficulties in capturing extreme weather. While post-processing techniques aim to enhance forecast accuracy, they rarely focus on precipitation, which exhibits complex spatial dependencies and tail behavior. Our novel framework leverages graph neural networks to post-process ensemble forecasts, specifically modeling the extremes of the underlying distribution. This allows to capture spatial dependencies and improves forecast accuracy for extreme events, thus leading to more reliable forecasts and mitigating risks of extreme precipitation and flooding. \footnote{Our code is available on \repo{GitHub}.}
\end{abstract}

\section{Introduction}
The increasing impacts of climate change have led to more frequent and severe extreme precipitation events, posing significant risks to infrastructure, agriculture, and public safety \citep{trenberth}. Accurate and well-calibrated precipitation forecasts are critical for mitigating these risks, especially concerning important applications such as disaster management, urban planning, and water resource management \citep{ipcc2021summary}. Ensemble prediction systems (EPS) have become a cornerstone of modern weather forecasting, generating probabilistic predictions by running numerical weather predictions (NWPs) under varied initial conditions. Despite their widespread use, ensemble forecasts often exhibit biases, a lack of sharpness, and difficulty capturing the extreme tail behavior of precipitation distributions, which limits their utility for decision-making under the increased risks associated with climate change \citep{tabari, trenberth}.

\begin{center}
    \begin{figure}[htb]
    \centering
\includegraphics[width=\linewidth]{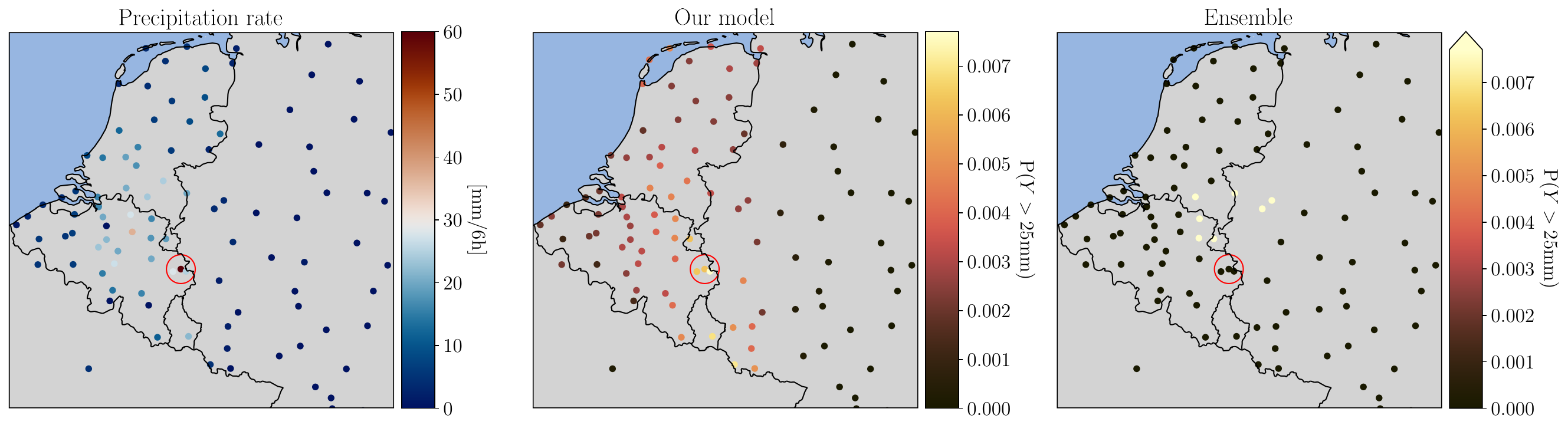}
    \caption{The left panel shows the precipitation rates on April 29, 2018 with a highlighted extreme precipitation occurrence of about 70mm over 6 hours. The two right plots show the threshold exceedance probability $\mathbb{P}[Y>25\text{mm}]$ for our model and the ensemble prediction. In contrast to the ensemble, our model assigns a nonzero probability ($0.63\%$) to the corresponding event.}
    \label{fig:spatial_extremes}
\end{figure}
\vspace{-5mm}
\end{center}

Post-processing techniques have been developed to address the limitations of raw ensemble forecasts by refining them into more accurate probabilistic predictions. Statistical methods such as ensemble model output statistics \citep{CalibratedProbabilisticForecastingUsingEnsembleModelOutputStatisticsandMinimumCRPSEstimation}, random forests \citep{npg-30-503-2023}, or nonparametric regression \citep{bremnes_2019} have been widely used to improve the calibration and sharpness of ensemble forecasts. More recently, neural-network-based post-processing has shown promise by leveraging machine learning to learn high-dimensional relationships directly from data \citep{NeuralNetworksforPostprocessingEnsembleWeatherForecasts, MachineLearningMethodsforPostprocessingEnsembleForecastsofWindGustsASystematicComparison}. These post-processing approaches have been extended to convolutional \citep{DeepLearningforPostprocessingGlobalProbabilisticForecastsonSubseasonalTimeScales} and graph neural networks \citep{feik2024graphneuralnetworksspatial} and to post-processing of neural network prediction systems \citep{bülte2024uncertaintyquantificationdatadrivenweather}. While these approaches are effective for general forecasting tasks, they often fail to capture the complex spatial dependencies and heavy-tailed characteristics of precipitation data, particularly during extreme weather events.

In this work, we introduce a novel framework for improving precipitation forecasts by post-processing ensemble predictions. Our method addresses key issues in forecasting extremes by explicitly accounting for the frequent occurrence of dry periods with a point mass and using a generalized Pareto distribution to capture the tail behavior associated with heavy rainfall. To enhance spatial accuracy, we employ graph neural networks (GNNs), which represent weather stations and ensemble forecasts with regards to their spatial dependence structure. This graph-based approach improves the model’s ability to identify patterns and dependencies in extreme events that often span across regions \citep{feik2024graphneuralnetworksspatial}. We validate our framework on a benchmark dataset for ensemble post-processing methods in medium-range weather forecasting.

\section{Data}
To compare to existing methods, we utilize the EUPPBench dataset, a benchmark dataset for ensemble post-processing \citep{essd-15-2635-2023}. The dataset comprises 122 weather stations across Europe and includes medium-range ensemble forecasts from the European Centre for Medium-Range Weather Forecasts (ECMWF) and corresponding station observations, spanning from 1997 to 2018. The dataset includes both, typical forecasts, but also reforecasts, which are numerical weather prediction (NWP) models run for past dates. In total, EUPPBench includes 730 daily operational forecasts with 51 ensemble members and 4180 reforecasts with 11 ensemble members and a total of 31 variables each \cite[compare][]{essd-15-2635-2023}. We follow the setup in \cite{feik2024graphneuralnetworksspatial}, where the model is trained on reforecast data from 1997-2013 and evaluated on reforecasts from 2014-2017, as well as on forecast data from 2017-2018. For modeling precipitation, we focus on predicting the \emph{TP6} variable, the total precipitation in $mm$ accumulated over six hours.

\section{Methodology}
We base our approach on a distributional regression network (DRN) \citep{NeuralNetworksforPostprocessingEnsembleWeatherForecasts}, a benchmark for station-based post-processing, that has been successfully applied to various domains, such as wind gusts \citep{MachineLearningMethodsforPostprocessingEnsembleForecastsofWindGustsASystematicComparison} or atmospheric rivers \citep{ProbabilisticPredictionsfromDeterministicAtmosphericRiverForecastswithDeepLearning}. The central idea is to use a neural network that outputs the parameters of a specified predictive distribution. The model is then trained by minimizing the Continuous Ranked Probability Score (CRPS), defined as $\text{CRPS}(F,y) \coloneq \int_{-\infty}^\infty (F(x) - \mathbbm{1}_{y \leq x})^2 \, dx$, where $F$ denotes the cumulative distribution function and $y$ denotes the realized outcome \citep{gneiting_katzfuss}. Similar to \cite{feik2024graphneuralnetworksspatial}, we employ a graph neural network to account for spatial dependencies and in addition choose a predictive distribution specifically designed to account for the characteristics of precipitation data.

\subsection{Probabilistic Precipitation modeling}
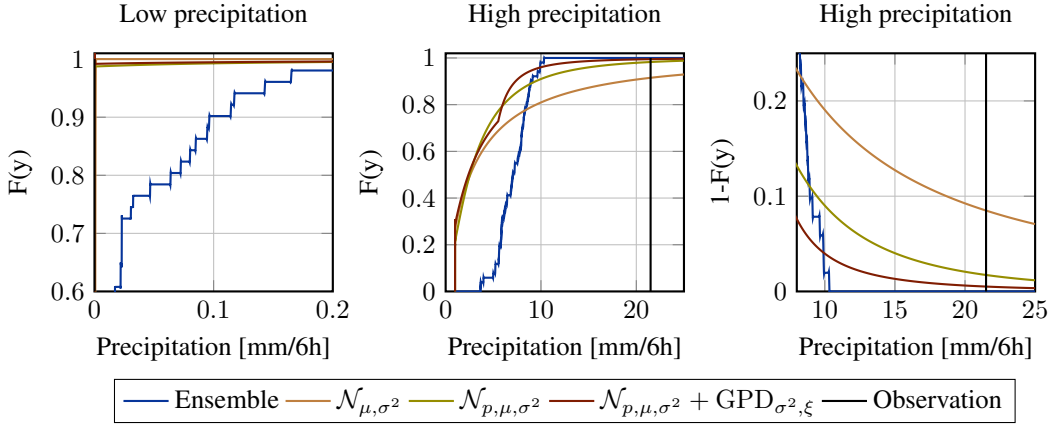
\begin{figure}[t!] 
    \centering
    \begin{tikzpicture}

\begin{groupplot}[
    group style = {group size=3 by 1, horizontal sep=1.5cm, vertical sep=1cm},
    width=4.75cm,
    height=4.75cm,
    xlabel={Precipitation [mm/6h]},
    grid=major,
    no markers,
    every axis plot/.append style={thick},
]

\nextgroupplot[
    title={Low precipitation},
    ymin=0.60, ymax=1.01, 
    xmin=0.0, xmax=0.20,  
    thick,
    smooth,
    ylabel={F(y)},
    xtick={0,0.1,0.2}
]
\coordinate (c1) at (rel axis cs:0,1);
\addplot[darkerblue] table[x=x, y=Ensemble, col sep=comma] {data/lp.csv};
\addplot[brown] table[x=x, y=Normal, col sep=comma] {data/lp.csv};
\addplot[olive] table[x=x, y={Normal Mixed}, col sep=comma] {data/lp.csv};
\addplot[darkerred] table[x=x, y=Mixed, col sep=comma] {data/lp.csv};
\addplot[black, ycomb, thick] coordinates {(0,1)};

\nextgroupplot[
    title={High precipitation},
    ymin=0.0, ymax=1.02, 
    xmin=0.0, xmax=25,
    thick,
    smooth,
    ylabel={F(y)}, 
]
\addplot[darkerblue] table[x=x, y=Ensemble, col sep=comma] {data/hp.csv};
\addplot[brown] table[x=x, y=Normal, col sep=comma] {data/hp.csv};
\addplot[olive] table[x=x, y={Normal Mixed}, col sep=comma] {data/hp.csv};
\addplot[darkerred] table[x=x, y=Mixed, col sep=comma] {data/hp.csv};
\addplot[black, ycomb, thick] coordinates {(21.5,1)};

\nextgroupplot[
    title={High precipitation},
    ymin=0.0, ymax=0.25, 
    xmin=8, xmax=25,
    thick,
    smooth,
    ylabel={1-F(y)},
    legend style={at={($(0,0)+(1cm,1cm)$)}, legend columns=5,fill=none,draw=black,anchor=center,align=center},
    legend to name=fred
]
\coordinate (c2) at (rel axis cs:1,1);
\addplot[darkerblue] table[x=x, y expr=1-\thisrow{Ensemble}, col sep=comma] {data/hp.csv};
\addplot[brown] table[x=x, y expr=1-\thisrow{Normal}, col sep=comma] {data/hp.csv};
\addplot[olive] table[x=x, y expr=1-\thisrow{Normal Mixed}, col sep=comma] {data/hp.csv};
\addplot[darkerred] table[x=x, y expr=1-\thisrow{Mixed}, col sep=comma] {data/hp.csv};
\addplot[black, thick] table[row sep = crcr]{21.5 0 \\ 21.5 0.3 \\};
\addlegendentry{Ensemble};    
\addlegendentry{$\mathcal{N}_{\mu, \sigma^2}$};    
\addlegendentry{$\mathcal{N}_{p, \mu, \sigma^2}$};  
\addlegendentry{$\mathcal{N}_{p, \mu, \sigma^2} + \mathrm{GPD}_{\sigma^2, \xi}$};    
\addlegendentry{Observation};  
\end{groupplot}
\coordinate (c3) at ($(c1)!.5!(c2)$);
\node[below] at (c3 |- current bounding box.south){\pgfplotslegendfromname{fred}};
\end{tikzpicture}
    \caption{The figure compares a sample prediction of the different modeling methods. In our proposed approach, the low precipitation scenarios are modeled via a discrete point mass that accounts for the event of no precipitation, while the GPD models the tail behavior over a certain threshold.}
    \label{fig:precip_predictions}
    \vspace{-5mm}
\end{figure}

While precipitation is one of the most critical meteorological variables, as its extremes have significant environmental and societal impacts, its highly variable nature and complex spatiotemporal characteristics make forecasting difficult. A dominant modeling approach in the literature is based on the mixed lognormal distribution \citep{article, cho2004}, which incorporates an additional point mass at zero with
probability $p$, accounting for instances of no precipitation.
A lognormal distribution describes a random variable $Y$ such that its logarithm, $\log(Y)$, follows a normal distribution.
While this has shown to perform well in practice \citep{article, syed}, the underlying data is heavily skewed, making neural network training difficult. Therefore, we propose to transform the underlying data space by applying a log-transform to the data, i.e., $h(y) = \log(y+\varepsilon)$, where we add $\varepsilon > 0$ for numerical stability (see \autoref{sec:appB} for more details). By definition, the transformed precipitation can now be modeled with a normal distribution with a discrete point mass at the left endpoint. As this distribution only accounts for the lower extreme of no precipitation, we further employ the Peaks-Over-Threshold approach to account for extreme high precipitation. By the Pickands-Balkema-De Haan theorem \citep{10.1214/aos/1176343003}, we know that under some regularity conditions, the excess function $\mathbb{P}(Y \leq x+u \mid Y>u)$ for a threshold $u$, converges to the generalized Pareto distribution (GPD). This allows to explicitly model the upper tails of the underlying distribution. To sum up, we model the (shifted and log-transformed) precipitation as
\begin{equation}
\label{eq:precip_cdf}
    F_Y(y) := \begin{cases}
        0, & y < c \\
         \Tilde{F}(y) \coloneq p + (1-p) \Phi_{\mu, \sigma^2}(y), & y \in [c,u]\\
         \Tilde{F}(u) + (1-\Tilde{F}(u)) \text{GPD}_{u.\sigma_u^2, \xi}(y), & y > u,        
    \end{cases}
\end{equation}
where $p$ denotes the discrete probability mass assigned to the left endpoint of the distribution $c \coloneqq \log(\varepsilon)$. We then use the neural network to predict the parameters $\{p,\mu,\sigma^2, \sigma_{u}^2, \xi, u\}$ per individual station with various meteorological variables from the NWP ensemble as input. The optimal parameters are obtained by minimizing the $\text{CRPS}(F_Y,y)$, for which a closed-form expression is given in \autoref{eq:full-crps} in \autoref{app:crps}.

\subsection{Spatial dependence modeling with Graph Neural Networks}
So far, the modeling has focused on station-specific features, with a single predictive distribution for each station. 
To apply graph-based learning methods, the data is transformed into a graph representation, where each station is treated as a node in the graph. Let $N$ denote the total number of stations, and define the distance matrix $D \in \mathbb{R}^{N \times N}$ based on geodesic distances between stations. An edge is created between nodes $i$ and $j$ if $D_{ij} \leq d_{\text{max}}$, where $d_{\text{max}}$ is a predefined distance threshold. 
To incorporate ensemble forecasts, every node is associated with a $n_{\text{ens}} \times F$-dimensional feature matrix, where $n_{\text{ens}}$ is the number of ensemble members and $F$ denotes the corresponding number of meteorological features. Edge weights $w_{i,j}$ between nodes $i$ and $j$ are defined based on normalized geodesic distances, capturing spatial relationships between stations. 

Since the ensemble members are interchangeable, the ensemble dimension introduces permutation symmetry. To account for this, the input features are embedded using a DeepSet \citep{zaheer2017deep}, ensuring permutation invariance. Specifically, we compute the initial node embedding as
$
h_v^{(0)} = \Psi\left(\sum_{n=1}^{n_{\text{ens}}} \rho\left(x_{v,n}\right)\right),
$
where $x_{v,n} \in \mathbb{R}^F$ is the node feature vector for station $v$ and ensemble member $n$. Here, $\rho$ and $\Psi$ are both two-layer multilayer perceptrons (MLPs). This setup allows the model to aggregate ensemble features while respecting their permutation-invariant nature. Unlike \citet{feik2024graphneuralnetworksspatial}, we perform this step before the GNN to reduce the dimensionality.
The GNN processes the graph with input node features $h_v^{(0)}$ by iteratively aggregating features from neighboring nodes. In addition, residual connections are applied to stabilize learning:
$
h^{(t)} = h^{(t-1)} + \sigma\left(\text{GNN}\left(h^{(t-1)}\right)\right),
$
where $h^{(t)}$ represents the hidden representation at layer $t$, and $\sigma$ is the ReLU activation function. Furthermore, we employ a Graph Isomorphism Network (GINE) \citep{xu2018how, Hu*2020Strategies}, which incorporates both node features and edge weights to effectively model interactions between neighboring nodes. After aggregation, the resulting features predict the station-specific parameters $\{p,\mu,\sigma^2, \sigma_{u}^2, \xi, u\}$. To ensure parameter constraints, we use a softplus activation for $\sigma^2, \sigma_u^2$, sigmoid activation for $p$ and $\xi$, and linear activation for $\mu, u$.

\section{Results}
We evaluate our approach on the EUPPBench dataset, comparing it against three baselines: ensemble prediction (ENS), a normal distribution ($\mathcal{N}_{\mu,\sigma^2}$), and a normal distribution with a point mass ($\mathcal{N}_{p,\mu,\sigma^2}$). Implemented in PyTorch with early stopping (compare \autoref{app:details}), our method uses two threshold selection strategies: (1) a global threshold, $u$, set as the 90th percentile of the training data ($\mathcal{N}$-$\text{GPD}_{\sigma_u^2}$); (2) and learned station-specific thresholds, $u_i$ ($\mathcal{N}$-$\text{GPD}_{u,\sigma_u^2}$). Due to numerical instability when optimizing the CRPS with respect to the GPD shape parameter, $\xi$, we fixed $\xi$ at 0.5 after a small hyperparameter search. Performance is evaluated using the CRPS, the Brier score for the binary event of no rain (= precipitation less than $0.01mm/6h$), and the quantile score (QS) at $\alpha = 0.99$ for extreme precipitation. These metrics collectively evaluate the entire predictive distribution, including both the lower and upper tails. 
\begin{table}
\centering
\begin{tabular}{llllllllll}
\toprule
Model & \multicolumn{3}{c}{24h} & \multicolumn{3}{c}{72}  & \multicolumn{3}{c}{120h}  \\
\midrule
Metric & CRPS & Brier & QS$_{0.99}$ & CRPS & Brier & QS$_{0.99}$ & CRPS & Brier & QS$_{0.99}$  \\
\midrule
ENS & 0.662 & 0.180 & 0.108 & 0.699 & 0.180 & 0.106 & 0.797 & 0.200 & 0.117 \\
$\mathcal{N}_{\mu,\sigma^2}$ & 0.515 & 0.316 & 0.299 & 0.640 & 0.384 & 0.381 & 0.782 & 0.337 & 0.558 \\
$\mathcal{N}_{p,\mu,\sigma^2}$ & \textbf{0.467} & \textbf{0.092} & \textbf{0.077} & \textbf{0.569} & \textbf{0.114 }& \textbf{0.092} & 0.682 & 0.139 & 0.117 \\
$\mathcal{N}$-$\text{GPD}_{\sigma_u^2}$ & 0.470 & 0.093 & 0.084 & 0.577 & 0.116 & 0.096 & \textbf{0.678} & 0.138 & \textbf{0.113} \\
$\mathcal{N}$-$\text{GPD}_{u,\sigma_u^2}$ & \textbf{0.467} & \textbf{0.092} & 0.082 & 0.597 & 0.119 & 0.099 & \textbf{0.678} & \textbf{0.137} & \textbf{0.113} \\
\bottomrule
\end{tabular}
    \caption{The table shows the evaluation metrics for the different models and some selected lead times on the forecasting task. The best model is highlighted in bold.}
    \label{tab:results_f}
\vspace{-5mm}
\end{table}

\autoref{tab:results_f} presents the results for the forecast task, with additional reforecast results available in the supplementary materials. The $\mathcal{N}_{p,\mu,\sigma^2}$ model performs well for most lead times, whereas the additional GPD modeling achieves similar performance and tends to outperform other methods at the 120h lead time. As the zero probability $p$ accounts for much of the data, we cannot expect the metrics to be too different. \autoref{fig:precip_predictions} visualizes the different probabilistic predictions for a high precipitation and zero precipitation event at a selected station, demonstrating a good fit between the predictive distribution and the realized event. In addition, \autoref{fig:spatial_extremes} shows a spatial visualization of a selected extreme precipitation event. In contrast to the ensemble prediction, our approach leads to a higher probability of the extreme event occurring. In addition, it correctly assigns low probability to sites that are spatially separated and where no precipitation occurred. Although more prominent for the lower tail, the above results suggest that our proposed approach can model the full range of precipitation, by explicitly considering the extremes on both ends of the support of the distribution.

\section{Conclusion}
We propose a predictive modeling framework for precipitation post-processing that focuses directly on the underlying extremes. By combining the modeling framework with a powerful graph neural network architecture, we can provide improvements in predictions regarding different baselines and with a focus on the extremes of the precipitation, allowing for more thorough prediction of extreme precipitation, mitigating risks of climate-change related events such as floods.
Possible future research might revolve around combining our modeling approach with an end-to-end neural network model, such as Graphcast \citep{graphcast} or RainNet \citep{rainnet}, to work with direct forecasting tasks. In addition, more detailed investigation of the extreme modeling with the GPD approach is required, especially with regards to the choice of the threshold $u$ and the shape parameter $\xi$. Further improving the tail modeling provides an interesting direction of research, regarding analysis of precipitation extremes.

\section*{Acknowledgements}
C. Bülte and G. Kutyniok acknowledge support by the DAAD programme Konrad Zuse Schools of Excellence in Artificial Intelligence, sponsored by the Federal Ministry of Education and Research.

S. Maskey acknowledges support by the NSF-Simons Research Collaboration on the Mathematical and Scientific
Foundations of Deep Learning (MoDL) (NSF DMS 2031985).

P. Scholl and G. Kutyniok acknowledge support by the project "Genius Robot" (01IS24083), funded by the Federal Ministry of Education and Research (BMBF), as well as the ONE Munich Strategy Forum (LMU Munich, TU Munich, and the Bavarian Ministery for Science and Art).

J. Berg and G. Kutyniok acknowledge support by the gAIn project, which is funded by the Bavarian Ministry of Science and the Arts (StMWK Bayern) and the Saxon Ministry for Science, Culture and Tourism (SMWK Sachsen).

G. Kutyniok acknowledges partial support by the Munich Center for Machine Learning (BMBF), as well as the German Research Foundation under Grants DFG-SPP-2298, KU 1446/31-1 and KU 1446/32-1. Furthermore, G. Kutyniok is supported by LMUexcellent, funded by the Federal Ministry of Education and Research (BMBF) and the Free State of Bavaria under the Excellence Strategy of the Federal Government and the Länder as well as by the Hightech Agenda Bavaria.


\bibliography{references}
\bibliographystyle{iclr2024_conference}

\appendix

\clearpage
\section{Additional results}
\begin{table}[htb]
\centering
\begin{tabular}{llllllllll}
\toprule
Model & \multicolumn{3}{c}{24h} & \multicolumn{3}{c}{72}  & \multicolumn{3}{c}{120h}  \\
\midrule
Metric & CRPS & Brier & QS$_{0.99}$ & CRPS & Brier & QS$_{0.99}$ & CRPS & Brier & QS$_{0.99}$  \\
\midrule
ENS & 0.757 & 0.196 & 0.208 & 0.854 & 0.216 & 0.221 & 0.965 & 0.233 & 0.270 \\
$\mathcal{N}_{\mu,\sigma^2}$  & 0.585 & 0.319 & 0.376 & 0.735 & 0.385 & 0.507 & 0.915 & 0.341 & 0.804 \\
$\mathcal{N}_{p,\mu,\sigma^2}$  mixed & \textbf{0.523} & 0.103 & 0.089 & \textbf{0.652} & \textbf{0.131} & 0.105 & \textbf{0.782} & 0.158 & 0.120 \\
$\mathcal{N}$-$\text{GPD}_{\sigma_u^2}$ & 0.530 & 0.104 & \textbf{0.095 }& 0.656 & 0.132 & \textbf{0.102} & 0.786 & 0.159 & 0.133 \\
$\mathcal{N}$-$\text{GPD}_{u,\sigma_u^2}$ & 0.524 & \textbf{0.102} & 0.096 & 0.676 & 0.136 & 0.117 & \textbf{0.782} & \textbf{0.157} & \textbf{0.116} \\
\bottomrule
\end{tabular}
    \caption{The table shows the evaluation metrics for the different models and some selected lead times on the reforecasting task. The best model is highlighted in bold.}
    \label{tab:results_rf}
\end{table}

\section{Precipitation Modeling}
\label{sec:appB}
Let $Y$ denote the precipitation variable in the unit $[mm/6h]$. To simplify the training procedure, we apply two successive transformations to the data. First, we shift our data by a small constant value $\varepsilon >  0$\footnote{ For our experiments we chose $\varepsilon = 0.01$.}. We model this shifted variable with a mixture distribution of the following form:
\begin{equation}
\label{eq:log_normal_point_mass}
    F_Y(y) \coloneq p + (1-p) \text{LN}_{\mu, \sigma^2}(y), \quad \varepsilon \leq  y.
\end{equation}
Here, $p \in [0,1]$ is the probability of the point mass at $\varepsilon$ and $\text{LN}$ denotes the log-normal distribution, defined as
\begin{equation}
    \text{LN}_{\mu, \sigma^2}(x)  \coloneq 
 \mathcal{N}_{\mu, \sigma^2}(\log(x))  =\Phi\left(\frac{\log(x) - \mu}{\sigma} \right).
\end{equation}
with $\Phi$ the cumulative distribution function (CDF) of a centered normal distribution. 
The combined distribution in \autoref{eq:log_normal_point_mass} is also known as the mixed lognormal distribution \citep{article} and has been commonly applied to precipitation modeling \citep{article, cho2004}.

\paragraph{Log-transform}
To remove skewness from the data, we perform a second transformation  $\Tilde{Y} = g(Y) \coloneq \log(Y)$, $Y \sim F_Y$. Note, for numerical stability, it is crucial that we have introduced the threshold $\varepsilon > 0$, thereby avoiding the singularity at $\log(0)$. We now want to analyze the resulting distribution $F_{\Tilde{Y}}(y)$. Using the cumulative distribution function (CDF), we obtain:

\[
F_{\Tilde{Y}}(y) = P \{ \Tilde{Y} \leq y \} = P \{ g(Y) \leq y \} = P \{ X \leq g^{-1}(y) \} = F_Y(g^{-1}(y)).
\]
Since $g^{-1}(y) = e^y$, we obtain for our transformed distribution:
\begin{equation}
    F_{\Tilde{Y}}(y) \coloneq p + (1-p) \mathcal{N}_{\mu, \sigma^2}(y) \quad  \log(\varepsilon) \leq y.
\end{equation}

\paragraph{Modeling Pareto in transformed space}
To model the upper tails of the distribution, we utilize the Pickands-Balkema-de Haan theorem \citep{10.1214/aos/1176343003}, which states that, under some regularity assumptions, the excess distribution of a random variable $\mathbb{P}(\Tilde{Y}\leq x+u \mid \Tilde{Y} > u )$ for a certain threshold $u$ can be approximated by a generalized Pareto distribution ($\text{GPD}_{u, \sigma_u^2,\xi}$), defined as
\begin{equation}
        \text{GPD}_{u,\sigma_u, \xi}(x)  \coloneq \begin{cases}
        1 - \left(1+ \frac{\xi (x-\mu)}{\sigma} \right)^{-1 / \xi}, & \xi \neq 0, \\
        1-\exp \left( -\frac{x-\mu}{\sigma} \right), & \xi = 0.
    \end{cases}
\end{equation}
For our modeling, this leads to the final definition of the CDF as
\begin{equation}
\label{eq:cdf_transformed}
    F_{\Tilde{Y}}(y) \coloneq  \begin{cases}
        \Tilde{F}(y) \coloneq p + (1-p) \mathcal{N}_{\mu, \sigma^2}(y), &  \log(\varepsilon) \leq y \leq u, \\
        \Tilde{F}(u) + (1 - \Tilde{F}(u)) \text{GPD}_{u,\sigma_u, \xi}(y), & y > u. \end{cases}
\end{equation}

\clearpage
\section{CRPS of the mixture distribution}
\label{app:crps}

As a loss function, we want to utilize the continuous ranked probability score (CRPS), which is a proper scoring rule and, therefore, can be used to measure the distance between a predictive distribution and a single data point \cite{gneiting_katzfuss}. It is defined as 

\begin{equation}
    \text{CRPS}(F,y)\coloneq \int_{-\infty}^\infty (F(x) - \mathbbm{1}_{y \leq x})^2 \, dx = \int_{-\infty}^yF^2(x)dx+\int_y^{\infty}(1-F(x))^2dx.
\end{equation}

for a given CDF $F$ and observation $y$. To compute it efficiently during training, it is important to compute the closed-form CRPS for the predictive distribution, shown in Equation~\ref{eq:cdf_transformed}.




For the ease of computation, we rewrite $F_{\hat{Y}}(y)$ as
\begin{equation}
    F_{\hat{y}}(y) = \begin{cases}
        F_1(y), & y \leq u \\
        F_2(y), & y > u,\\
    \end{cases}
\end{equation}
where we assume that $y\geq c \coloneqq\log(\epsilon)$. Furthermore, we denote
\begin{equation}
    F_1(y) = \begin{cases}
        0, & y < c \\
        p + (1-p) \Phi_{\mu, \sigma^2} (y), & y \in [c, u)\\
        1, & y \geq u,
    \end{cases}
\end{equation}
where $ \Phi_{\mu, \sigma^2} (y)$ denotes the CDF of the normal distribution with mean $\mu$ and standard deviation $\sigma$,
and
\begin{equation}
    F_2(y) = \begin{cases}
        0, & y < u \\
        F_1(u) + (1-F_1(u)) \text{GPD}_{u, \sigma_u, \xi}(y), & y \geq u.
    \end{cases}
\end{equation}
Then, following \cite{Jordan2016_1000063629} we can decompose the CRPS into the following form:
\begin{equation} \label{eq:full-crps}
    \text{CRPS}(F_{\hat{y}}, y) = \begin{cases}
        \text{CRPS}(F_1, y) + \text{CRPS}(F_2, u), & y < u \\
        \text{CRPS}(F_1, u) + \text{CRPS}(F_2, y), & y \geq u.
    \end{cases}
\end{equation}

\autoref{eq:full-crps} gives the loss of the predicted CDF proposed in this paper: a mixture of a truncated normal distribution with point mass and GPD for extreme values. The different parts of it are computed in Equation~\ref{eq:CRPS-F2-y}, \ref{eq:CRPS-F2-u}, \ref{eq:CRPS-F1-y}, and \ref{eq:CRPS-F1-u}.

\paragraph{CRPS for $F_2$}
\cite{jordan2018evaluatingprobabilisticforecastsscoringrules} provide a closed-form expression for the Generalized Pareto distribution with point mass, which is given as
\begin{equation} \label{eq:CRPS-F2-y}
    \text{CRPS}(F_2, y) = \sigma_u \left( \frac{y-u}{\sigma_u} - \frac{2(1-M)}{1-\xi} \left( 1- \left(1- F_\xi\left(\frac{y-u}{\sigma_u}\right)\right)^{1-\xi} \right) + \frac{(1-M)^2}{2-\xi} \right),
\end{equation}
where in our case $M = F_1(u)$. In addition, we obtain
\begin{equation} \label{eq:CRPS-F2-u}
    \text{CRPS}(F_2,u) = \sigma_u \frac{(1-M)^2}{2-\xi}.
\end{equation}
Note that the CRPS for the GPD distribution is only defined for $\xi < 1$.

\paragraph{CRPS for $F_1$}
We use the following representation of the CRPS by \cite{Jordan2016_1000063629}, which we first derive for the case of a standard normal distribution. Denote $F_{c,\mu, \sigma^2}^u \coloneq F_1$, where $\mu$ and $\sigma$ are the mean of the normal distribution. Then we have
\begin{equation}  \label{eq:CRPS-F1-y}
\begin{aligned}
    \text{CRPS}(F_{c,0, 1}^u, y) &=  \, y(2F_{c,0, 1}^u(y) - 1) - cP_c^2 + uP_u^2 + 2(1-p)G(c)P_c + 2(1-p)G(u)P_u \\
    & - 2
    \begin{cases} 
        (1-p)G(c) - cP_c, & y < c, \\
        (1-p)G(y), & c \leq y < u, \\
        (1-p)G(u) +uP_u, & y \geq u,
    \end{cases} \\
    & + 2(1-p)^2 \int_c^u G(x)f(x) \, dx,
\end{aligned}
\end{equation}
where $f=\varphi$, $P_x = F_1(x) - F_1(x^-)$, and $G(x) = \int_{- \infty}^x t \varphi(t) \, dt$ with $\varphi$ denoting the density of a standard normal distribution. First, note that $G(x) = -\varphi(x)$. Then, we can consider each term individually:
\begin{align*}
    P_c &= p + (1-p)\Phi(c) \\
    P_u &= 1 - (p + (1-p)\Phi(u)) = (1-p)(1-\Phi(u))\\
    G(c) &= - \varphi(c) \\
    G(u) &= - \varphi(u)\\
    G(y) &= - \varphi(y)\\
\end{align*}
The only remaining unknown term is the last integral, which can be expressed as
\begin{align}
    \int_c^u G(x)f(x) \, dx &= - \int_c^u \varphi^2 \, dx = - \frac{1}{2\sqrt{\pi}}\left(\Phi(\sqrt{2}u) - \Phi(\sqrt{2}c) \right)
\end{align}
Following \cite{jordan2018evaluatingprobabilisticforecastsscoringrules}, we know that for a location-scale transformation, we have
\begin{equation}
    \text{CRPS}(F_1,y) = \sigma \text{CRPS}\left(F_{(c-\mu)/\sigma,0, 1}^{(u-\mu) / \sigma}, \frac{y-\mu}{\sigma}\right) 
\end{equation}
Lastly, we can compute
\begin{equation} \label{eq:CRPS-F1-u}
    \begin{aligned}
    \text{CRPS}(F_{c,0, 1}^u,u) &= u - c P_c^2 + uP_u^2 + 2(1-p)G(c)P_c + 2(1-p)G(u)P_u\\
    &- 2 ((1-p)G(u) + u P_u) - (1-p)^2  \frac{1}{\sqrt{\pi}}\left(\Phi(\sqrt{2}u) - \Phi(\sqrt{2}c) \right)
\end{aligned}
\end{equation}

\section{Experimental Details}
\label{app:details}
In this section, we describe the experimental setup in detail, covering the software libraries used, the graph construction process, the permutation-invariant ensemble embedding, and the specifics of our Graph Neural Network (GNN) architecture, including the Graph Isomorphism Network with Edge features (GINE). All experiments were implemented in PyTorch \citep{NEURIPS2019_9015} and PyTorch Geometric (PyG) \citep{Fey/Lenssen/2019}. The code for our method will be made public upon acceptance.

\subsection*{Graph Construction.}
Meteorological stations are modeled as nodes in a graph. For $N$ stations, we first compute the geodesic distance matrix $D \in \mathbb{R}^{N \times N}$, where each element $D_{u,v}$ represents the geodesic distance between station $u$ and station $v$. An edge is created between nodes $u$ and $v$ if 
\[
D_{u,v} \leq d_{\max},
\]
with $d_{\max} = 300\,\text{km}$ in our experiments. Each node is assigned a feature matrix of dimensions $n_{\text{ens}} \times F$, where $n_{\text{ens}}$ is the number of ensemble members and $F$ denotes the number of meteorological features.

Edge weights $w_{u,v}$ are computed based on the inverse of the geodesic distance between the corresponding stations. Specifically, given the locations $l_u$ and $l_v$ for stations $u$ and $v$, respectively, the weight is defined as:
\[
w_{u,v} = \frac{1}{d(l_u, l_v)},
\]
where $d(l_u, l_v)$ is the normalized geodesic distance between the two locations. After computing these weights, we normalize them and set the self-connection weight $w_{u,u} = 1$ for all nodes $u$.

We summarize the graph as $G = (V, E, X, D, W)$, where $V$ is the set of nodes, $E$ is the set of edges, $X$ represents the node feature matrices, $D$ is the distance matrix, and $W$ is the weight matrix.

\subsection*{Permutation-Invariant Ensemble Embedding}

To address the permutation symmetry in ensemble forecasts, where the order of ensemble members is irrelevant, we incorporate a DeepSet architecture \citep{zaheer2017deep} into the preprocessing stage. The initial node embedding for station $v$ is computed as:
\[
h_v^{(0)} = \Psi\!\left(\sum_{n=1}^{n_{\text{ens}}} \rho\!\left(x_{v,n}\right)\right),
\]
where $x_{v,n} \in \mathbb{R}^F$ is the feature vector corresponding to ensemble member $n$ at node $v$. The functions $\rho$ and $\Psi$ are implemented as two-layer multilayer perceptrons (MLPs), ensuring that the embedding remains invariant to the permutation of ensemble members.

\subsection*{Graph Neural Network Architecture}

The core of our model is a Graph Neural Network with residual connection that processes the constructed graph to capture spatial dependencies. Specifically, we use a Graph Isomorphism Network with Edge features (GINE) \citep{xu2018how,Hu*2020Strategies} to integrate both node and edge information.

For each node $v$ at layer $t$, the GINE update the node feature via:
\[
h_v^{(t)} = h_v^{(t-1)} +\operatorname{MLP}^{(t)}\!\left((1+\epsilon^{(t)}) \cdot h_v^{(t-1)} + \sum_{u \in \mathcal{N}(v)} \operatorname{ReLU}\Bigl(h_u^{(t-1)} + w_{u,v}\Bigr)\right),
\]
where $\mathcal{N}(v) := \{ u \in V \, | \, (v,u) \in E \}$.

\subsection*{Output Layer and Parameter Prediction}

The final node representations are used to predict station-specific parameters $\{p, \mu, \sigma^2, \sigma_u^2, \xi, u\}$. To enforce appropriate parameter constraints, we apply:
\begin{itemize}
    \item a \emph{softplus} activation for $\sigma^2$ and $\sigma_u^2$ and
    \item a \emph{sigmoid} activation for $p$ and $\xi$ and
    \item a \emph{linear} activation for $\mu$ and $u$.
\end{itemize}

\subsection*{Optimization and Training}

Our model is trained using the Adam optimizer \citep{kingma2017adammethodstochasticoptimization} with a learning rate set to 0.0001. Training and evaluation are performed on NVIDIA Tesla RTX3090 GPUs with 24GB RAM. Each model is trained for 25 epochs, and we report the test CRPS at the epoch that achieves the lowest validation CRPS.

\end{document}

%% file: math_commands.tex
\usepackage{amsmath,amsfonts,bm}

\def\eqref#1{equation~\ref{#1}}

\def\1{\bm{1}}

\DeclareMathAlphabet{\mathsfit}{\encodingdefault}{\sfdefault}{m}{sl}
\SetMathAlphabet{\mathsfit}{bold}{\encodingdefault}{\sfdefault}{bx}{n}

